\documentclass[runningheads]{llncs}
\usepackage{graphicx}
\usepackage{amsmath,amssymb} 
\usepackage{color}
\usepackage{multirow}
\usepackage{subfigure}
\usepackage[export]{adjustbox} 
\usepackage{cite}
\usepackage{booktabs} 
\usepackage[width=122mm,left=12mm,paperwidth=146mm,height=193mm,top=12mm,paperheight=217mm]{geometry}
\newcommand{\thirdsplitwidth}{0.32}

\begin{document}

\title{Learning to Count Objects \\ with Few Exemplar Annotations} 



\author{Jianfeng Wang \and
	Rong Xiao \and
	Yandong Guo \and
	Lei Zhang}
	
\institute{Microsoft \\
\email{\{jianfw, rxiao, yag, leizhang\}@microsoft.com}
}

\maketitle

\begin{abstract}
In this paper, we study the problem of object counting with incomplete annotations.
Based on the observation that in many object counting problems the target objects are normally repeated and highly similar to each other, we are particularly interested in the setting when only a few exemplar annotations are provided. Directly applying object detection with incomplete annotations will result in severe accuracy degradation due to its improper handling of unlabeled object instances. To address the problem, we propose a positiveness-focused object detector (PFOD) to progressively propagate the incomplete labels before applying the general object detection algorithm. The PFOD focuses on the positive samples and ignore the negative instances at most of the learning time. This strategy, though simple, dramatically boosts the object counting accuracy.
On the CARPK dataset for parking lot car counting, we 
improved mAP@0.5 from 4.58\% to 72.44\% using only 5 training images each with 5 bounding boxes. 
On the Drink35 dataset for shelf product counting, the mAP@0.5
is improved from 14.16\% to 53.73\% using 10 training images each with 5 bounding boxes. 

\end{abstract}

\section{Introduction}
Object counting is to count the number of object instances 
in a single image or video sequence. 
It has many real-world applications such as 
traffic flow monitoring, crowdedness estimation, and product counting.  

Existing approaches towards the counting problem can be roughly categorized as regression-based approach~\cite{AnLV07,ChanLV08,ChenGXL13,ChenLGX12,KongGT06}, density-based approach~\cite{ZhangLWY15,ArtetaLNZ14, RodriguezLSA11} and detection-based approach~\cite{HsiehLH17,KamenetskyS15,MoranduzzoM14}. 
The regression-based approach directly learns a mapping from the image to the 
number of instances. In contrast, the density-based approach first estimates a 
density map and then aggregates the density information to get the number of instances.
Both categories of approaches provide little information of exact instance location, whereas the 
detection-based approach is capable of detecting the location of 
each individual instance and thus making the counting result more explainable. Due to 
this advantage, we mainly focus on the detection-based approach in this work. 

\begin{figure}[t!]
\centering
\begin{tabular}{c@{~}c@{~}c}
\includegraphics[width=\thirdsplitwidth\linewidth]{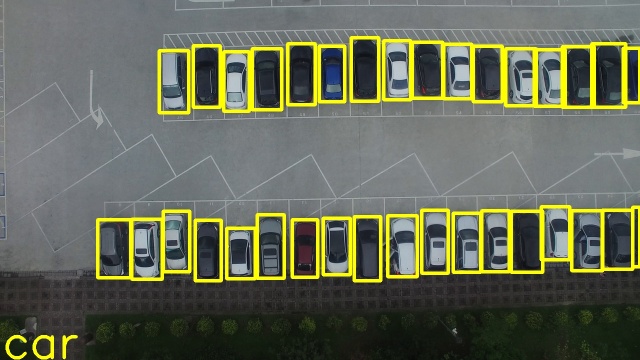} & 
\includegraphics[width=\thirdsplitwidth\linewidth]{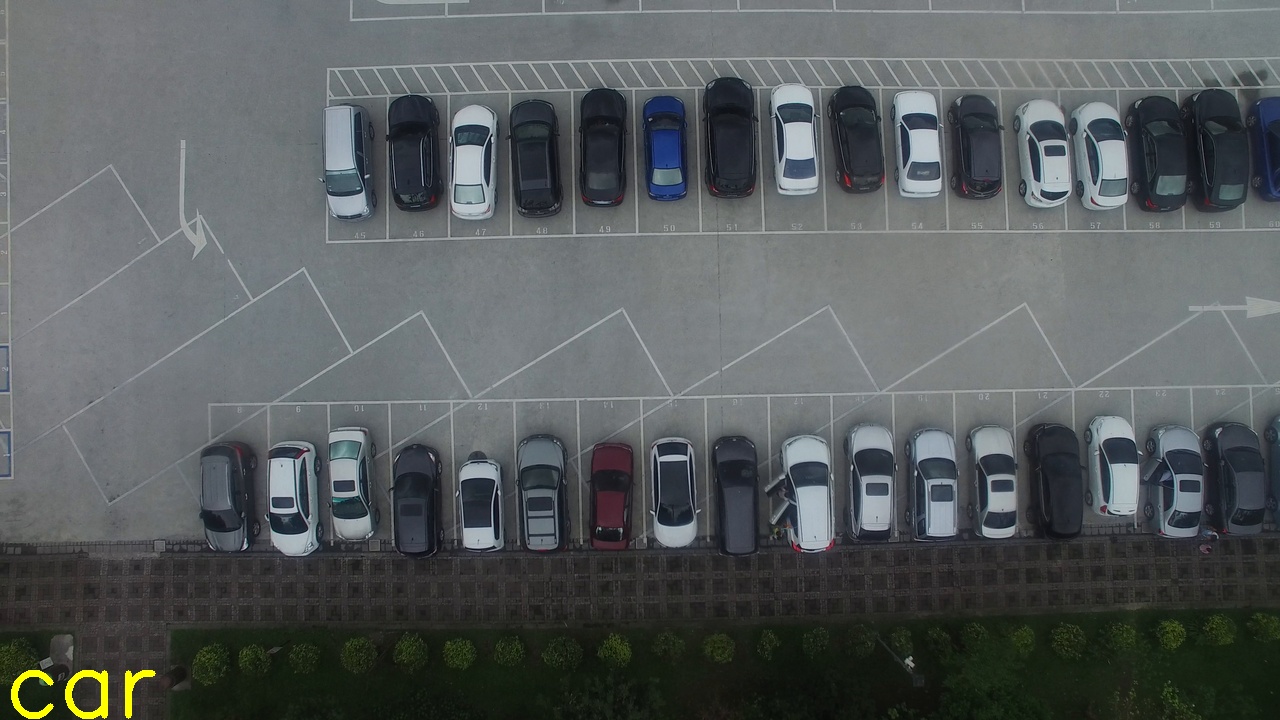} &
\includegraphics[width=\thirdsplitwidth\linewidth]{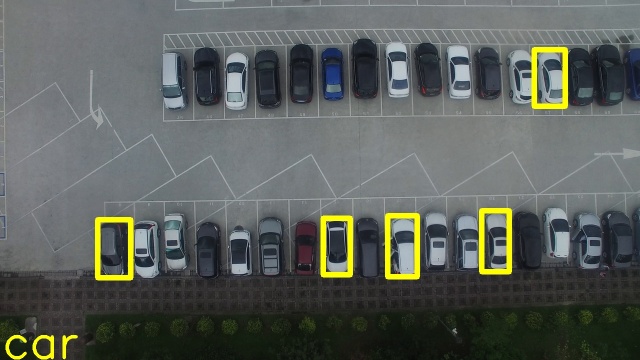} \\
\includegraphics[width=\thirdsplitwidth\linewidth]{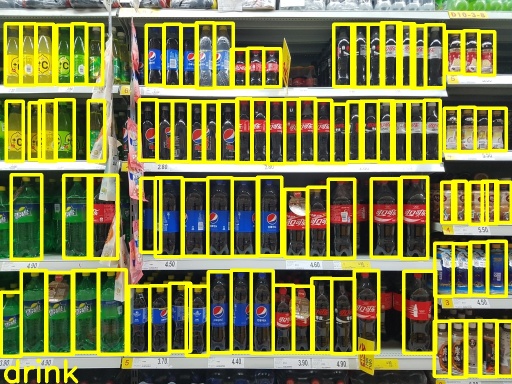} & 
\includegraphics[width=\thirdsplitwidth\linewidth]{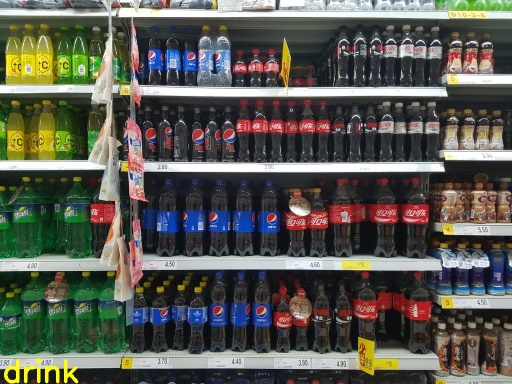} & 
\includegraphics[width=\thirdsplitwidth\linewidth]{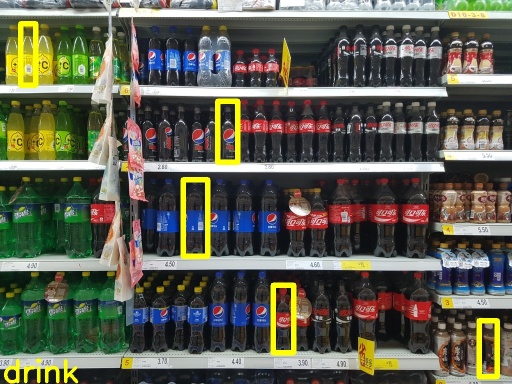} \\
(a) Fully-Supervised & 
(b) Weakly-Supervised & 
(c) Incompletely-Supervised 
\end{tabular}
\caption{Different training image settings in detection-based object counting. 
The label names are shown in bottom left
}
\label{fig:isl}
\end{figure}

In recent years, the accuracy of object detection has been dramatically improved~\cite{Girshick15,RenHGS15,LiuAESRFB16,RedmonF17} thanks to the advance of deep convolutional neural network. 
However, the high accuracy is achieved normally with the large number of fully annotated bounding boxes. 
In the context of object counting, the number of instances even in a single image 
could be huge (e.g. from tens to hundreds), which not only presents a great challenge to object detection, but also requires tremendous annotation effort to build a high-accuracy object
detector. 

To save the annotation cost, 
weakly supervised approaches
are generally adopted to train the model 
solely based on image-level annotations. However, such approaches still
suffer from suboptimal accuracy due to the lack of instance-level (e.g. bounding box) annotations.
To address this problem, we resort to a problem setting where each image is only partially annotated
with a few bounding boxes and leaves other instances unlabeled. This setting is practically useful especially when the number of instances is large and it is tedious and costly to label all the instances. 
Another motivation is that in certain object counting problem, the large number of instances exhibits less variances, e.g. scale, color. 
Potentially, a few annotated bounding boxes might generalize well to achieve a high accuracy.  
We call the learning algorithm trained on such training data \textit{incompletely-supervised learning}. 

Fig.~\ref{fig:isl} shows examples to illustrate the training image difference among the fully-supervised, weakly-supervised, and incompletely-supervised learning. 
For the fully-supervised setting, 
all cars and drinks are annotated.
The total number of the car instances is as many as 58. 
In the weakly-supervised setting, we only have the information that the car or drinks exist in the image. 
Based on the observation that in many object counting problems the target objects are normally repeated and highly similar to each other, we are particularly interested in the setting when only a few exemplar annotations are provided, which is illustrated in Fig.~\ref{fig:isl}(c)

In incompletely-supervised learning, if any unlabeled region is simply treated as background in object detection training,  
as the number of labeled instances is usually small, 
the trained model could over-fit the training data severely, and 
mistaken the unlabeled object as background. 
To address the issue, we propose a simple yet effective algorithm, named \textit{positiveness-focused object detector} (PFOD), to progressively propagate the labels, which treats unlabeled regions as background first and then neglects them to mainly focus on the positive samples during the training.
The intuition is to first learn a compact classifier which might over-fit the data, and then relax the learning by ignoring the unlabeled region to pull the unlabeled instances towards being positive. 
In this way, bounding box supervision could be automatically expanded during training
before we apply a standard object detector.


Overall, our contributions can be summarized as follows. 
\begin{enumerate}
\item We explicitly formulate the problem of incompletely-supervised learning, which focuses on the incomplete annotations for object counting. 
\item We propose a progressive label propagation algorithm through positiveness-focused object detector to properly handle the incomplete labels. 
\item We conduct extensive experiments demonstrates the effectiveness of the proposed PFOD to handle such training problems. 
On the CARPK dataset for parking lot car counting, we 
improved mAP@0.5 from 4.58\% to 72.44\% using only 5 training images each with 5 bounding boxes. 
On the Drink35 dataset for shelf product counting, the mAP@0.5
is improved from 14.16\% to 53.73\% using 10 training images each with 5 bounding boxes. 

\end{enumerate}

\section{Related Work}

\subsection{Object Counting}
The approaches towards object counting can be roughly categorized into three
categories: regression-based approach, density-based approach, and detection-based 
approach. 


\subsubsection{Regression-based Approach} predicts 
instance count directly based on global regressors with image features~\cite{ChanLV08, ChenGXL13, ChenLGX12, KongGT06}.
For example, in~\cite{ChanLV08}, the video sequence is first segmented into 
different components of homogeneous motion, and a Gaussian process regression is learned
for each segmented region to count the number of instances.
Cumulative attribute representation~\cite{ChenGXL13}
was proposed and used to learn the regressor to handle the imbalanced training data.
Inter-dependent features are used to mine the spatially importance among different region
to learn the number of count~\cite{ChenLGX12}. 
Feature normalization is taken into account to deal with perspective projections in~\cite{KongGT06}. With the advantage of the deep learning, ~\cite{MundhenkKSB16} 
estimate the number of cars using CNNs. 
These regression-based approaches provide no clue of individual location of each
object, which limits its potential applications. 


\subsubsection{Density-based Approach} 
first maps the image to a density map, such that 
the integral over
any sub region gives the count of objects within that region~\cite{LempitskyZ10, ArtetaLNZ14, RodriguezLSA11, ZhangLWY15}.
In~\cite{LempitskyZ10}, the pixel-level density is learned by minimizing a regularized risk quadratic cost function. 
Based on the density map, an interactive counting system is introduced 
in~\cite{ArtetaLNZ14} to incorporate the relevance feedback. 
Instead of hand-crafted feature, \cite{ZhangLWY15} focuses on the CNN-based 
density map and instance count estimation in cross-scene scenario. 
While the density map provides certain clue of the crowdedness, it still lacks the 
exact position of each instance.

\vspace{-0.2cm}
\subsubsection{Detection-based Approach} gives the total count by localizing each object instance.
Such approaches can be considered as the application of object detection and thus benefit a lot from the improvement of object 
detection~\cite{Girshick15,RenHGS15,LiuAESRFB16,RedmonF17}. However, directly applying object detection for object counting requires special focus on small-scale objects and extraordinary effort on massive instance labeling.
In~\cite{LinD10}, a hierarchical part-template matching approach is proposed to detection humans, which requires careful feature and template design, 
while in~\cite{HsiehLH17}, the neural network learning is 
applied to detect and count car instances by incorporating layout information. 
With its large potential in applications demanding location information, 
we mainly work on the detection-based approach with special focus on incompletely-supervised learning. 

\subsection{Object Detection}
Nowadays, mainstream object detection algorithms have changed to CNN-based implementation
 due to its powerful representation and high accuracy. Such algorithms can be roughly categorized into 
two-stage object detector~\cite{GirshickDDM14,Girshick15,RenHGS15},
and one-stage 
object detector~\cite{LiuAESRFB16,RedmonF17}.
Two-stage object detectors such as Faster RCNN~\cite{RenHGS15} first extract 
region proposals and then perform classification 
and bounding box regression, while 
single-stage objectors such as YOLO~\cite{RedmonF17} directly output the bounding box
locations and the classification results without generating the region proposal. 
All the object detection algorithms by default following the fully supervised setting - requiring 
large amount of training images with annotated bounding boxes to achieve a high accuracy. 

To reduce the annotation 
effort, weakly supervised learning
trains object 
detector only based on image-level labels. 
Most approaches~\cite{BilenV16, CinbisVS17, KantorovOCL16, TangWBL17} first generates 
multiple region proposals for each image and 
then leverage multi-instance
learning algorithms to solve the problem. However, such approaches still
suffer from suboptimal accuracy due to the lack of instance-level labels.

In both the fully and weakly supervised settings,
each image is either labeled with full bounding box annotations or 
only image-level category information. In contrast, we are more interested in  
the incomplete supervised setting, where
only a few exemplar bounding boxes are annotated
and all others are unlabeled.
This setting is practically more useful when the number of instances
is large, for example to count the number of products on a retail store shelf, or
the number of cars in a large parking lot.




\vspace{-0.2cm}
\section{Approach}
\vspace{-0.2cm}
\subsection{Problem Definition}
Although detection-based object counting resembles the general object detection problem, it still presents unique challenges 
when the total number of object instances in an image is large, e.g. from tens to hundreds, and the size of each instance is relatively small comparing with the image size.

To save the tedious and costly labeling effort, we assume each image $I_n$ in the training set $\{I_n|n=1,\ldots,N\}$ is only labeled with a few instances (e.g. less than ten) though there might be tens or hundreds of instances in one image. For image $I_n$, denote by $\mathbf{x}_i^{n}$ its $i$-th annotated bounding box ($i = 1, \ldots, M_1(n)$).
Since not all instances are annotated, the rest region could also contain the object instances. Without loss of generality, we assume the number of object categories is 1. For notation simplicity, we drop the $n$ from $\mathbf{x}_i^{n}$ and $M_1(n)$ when there is no ambiguity.
Thus, the problem is how to build an effective object detector to count the object instances based on the incomplete annotations $\{\mathbf{x}_i^n\}$. 
We name the problem as \textit{incompletely-supervised learning} for object counting. 

\subsubsection{Comparison with other supervised learning problems.}
The differences with other supervised learning settings in the context of object detection are also shown in Fig.~\ref{fig:isl}.
Fully-supervised learning 
	requires that all the bounding boxes are labeled. This represents the upper bound of detection-based object counting performance, but is costly to label every instance.
Weakly-supervised learning
	assumes that only image-level category information is available. That means, we know there are some object instances in one image, but we have no information of their locations. 
Another problem setting is low-shot learning, 
where the number of training images is small, but each training image has full annotations. 
This is more like the fully-supervised learning but with only a few training images.

\newcommand{\intuitionwidth}{0.105}
\begin{figure}[t!]
	\centering
	\begin{tabular}[t]{@{}c@{~~~~~}c@{~~~~~}c@{~~~~~}c@{~~~~~}c@{~~~~~}c@{}}
		\includegraphics[width=\intuitionwidth\linewidth,valign=T]{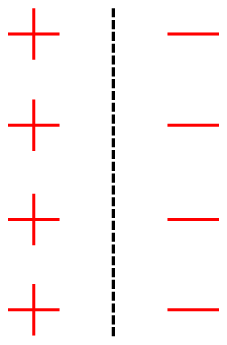}	&
		\includegraphics[width=\intuitionwidth\linewidth,valign=T]{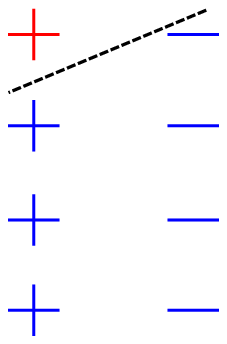} &
		\includegraphics[width=\intuitionwidth\linewidth,valign=T]{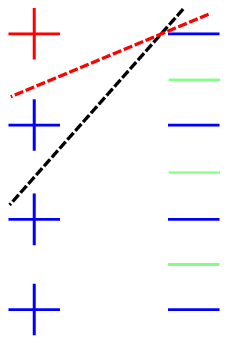} &
		\includegraphics[width=\intuitionwidth\linewidth,valign=T]{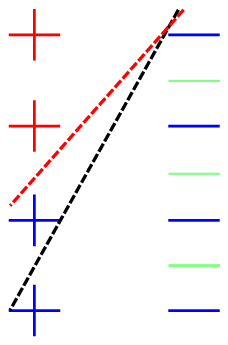} &
		\includegraphics[width=\intuitionwidth\linewidth,valign=T]{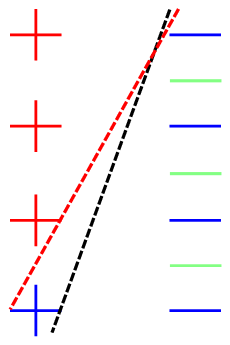} &
		\includegraphics[width=\intuitionwidth\linewidth,valign=T]{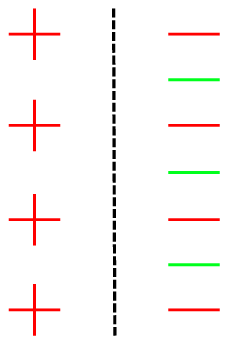} \\\\
		(a) Ideal & (b) Baseline & (c) Stage 1 & (d) Stage 2 
		& (d) Stage 3 & (e) Stage 4 \\ 
	\end{tabular}
	\caption{Intuition of the proposed label propagation through a positiveness-focused object detector to manage the incompletely-supervised labels}
	\label{fig:intuition}
\end{figure}

\subsection{Label Propagation by Positiveness-Focused Object Detection}
As we have a few exemplar labels of the target objects, the most intuitive way is to treat the labels as seeds and carefully propagate them to unknown regions. 
We propose a simple yet effective \textit{positiveness-focused object detector} (PFOD) to solve the propagation problem.

\subsubsection{Intuition.} Fig.~\ref{fig:intuition} illustrates the intuition of the label propagation by PFOD. 
The plus symbol denotes a positive sample while the minus symbol denotes negative.
The sample in red is known (labeled) while the sample in blue is unknown. 
The line with the dotted line is the decision boundary. 
If all the positive and negative samples are known, we can easily figure out the decision boundary
to separate true positive and true negative samples, as shown in Fig.~\ref{fig:intuition}(a).
When only one positive sample is known and all the rest are unknown, as shown in 
Fig.~\ref{fig:intuition}(b), if we simply treat all the unknowns as negative sample, the decision boundary
could mis-classify the unlabeled positive samples as being negative.

To compensate the lack of negative samples, we can introduce some images as extra negative data, as shown in green from Fig.~\ref{fig:intuition}(c) to 
Fig.~\ref{fig:intuition}(e). For object detection, it is indeed easy to find images without any target objects in the same domain.  
For example, we can use the PASCAL VOC~\cite{EveringhamGWWZ10} 2007 data set 
as extra background images for the problem of car counting on the CARPK~\cite{HsiehLH17} dataset. 

In Fig.~\ref{fig:intuition}(c) we show how to propagate the labels. We first treat all the unlabeled samples (shown as both blue plus and minus symbols in Fig.~\ref{fig:intuition}(c)) as negative samples 
to train the detector, and the learned decision boundary will be pushed close to the only positive sample (red plus symbol) as depicted with the red dotted line.
Next, we ignore the unknown samples (note that we still have the extra negative samples shown as green minus symbols) and gradually update the learned classifier to classify the labeled positive sample and the extra negative samples. As the unknown positive samples (shown as blue plus symbols) are not taken as negative samples, they no longer push the decision boundary. As a result, the decision boundary (as shown as black dotted line) will be moved a bit further from the known positive sample and classify a few more unknown positive samples as positive samples.

If the propagation is carefully controlled, we can treat the newly classified positive samples as known labeled data for the next stage, as shown in Fig.~\ref{fig:intuition}(d). We repeat the process above 
to iteratively learn the boundary and propagate the label set. Each iteration here is called 
a stage, and we use $L^{s}=\{\mathbf{x}_i|i=1,\ldots M_s\}$ to denote the label set in the $s$-th stage, where $M_s$ is the number of expanded bounding boxes in one image. Finally, we 
combine all the expanded positive samples and take all the others as background to learn the final 
decision boundary. 
 
\begin{figure}[t!]
	\centering
	\includegraphics[width=0.99\linewidth]{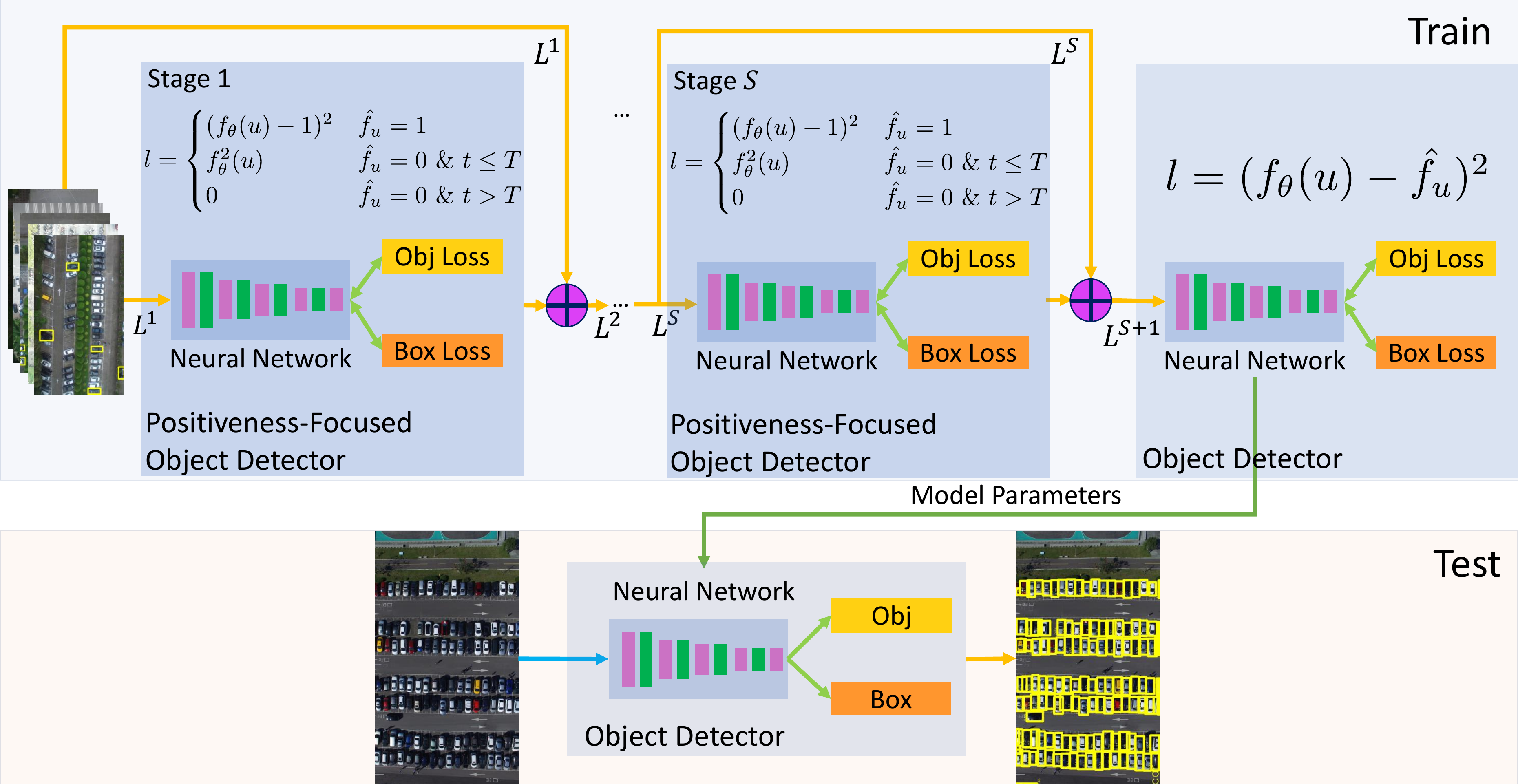}
	\caption{Framework of the proposed label propagation by a positiveness-focused object detector (PFOD)  
}
	\label{fig:framework}
\end{figure}

\subsubsection{Solution.} Fig.~\ref{fig:framework} shows the framework of the proposed strategy to learn the object detector. 
At the initial state of training, the bounding boxes we have are the labeled set, i.e. $L^1 = \{\mathbf{x}_i | i= 1, \ldots, M_1\}$ for each training image. 
Based on $L^s$, we train a positiveness-focused object detector (PFOD), 
which can be based on 
any object detector~\cite{GirshickDDM14,Girshick15,RenHGS15,LiuAESRFB16,RedmonF17}. 
In this work, we choose YoloV2~\cite{RedmonF17} for its simplicity. 
The network first processes each training image in a batch manner by a fully convolutional neural network, 
and then outputs three components at each spatial position: bounding box coordinates, objectiveness to tell how confident the bounding box contains an object, and classification scores to tell which category the bounding box contains. 
Here we assume the number of categories is 1 and remove the classification module. 
The Euclidean loss is used for bounding box coordinate regression and 
objectiveness confidence regression.

Specifically, for the objectiveness at spatial position $u$, 
the loss is defined as
\begin{align}
	l = (f_\theta(u) - \hat{f_u})^2
\end{align}
where $\theta$ is the network parameter, learned iteratively through the 
mini-batch stochastic gradient descent (SGD)
algorithm, $f_\theta(u)$ is the objectiveness score at position $u$,
and $\hat{f_u}$ is 1 if it is identified as being positive for position $u$, and 0 otherwise based on the current label set. 
For the extra background images, the label is consistently set as 0 and the loss will be always enabled. For the training images in the target domain, we modify it as 
follows to implement PFOD,

\begin{align}
l = \begin{cases}
(f_\theta(u) - 1)^2  & \hat{f_u} = 1 \\
f_\theta^2(u) & \hat{f_u} = 0 \text{ \& }  t \le T \\
0 & \hat{f_u} = 0  \text{ \& } t > T
\end{cases}
\end{align}
where $t$ is the number of 
iterations in SGD, $T$ is a pre-defined parameter (200 in experiments) to determine how many iterations are needed to treat the unknown regions as background.  
After $T$ iterations, the detector training will only focus on the positive samples and the extra background data. 

With the model trained by PFOD, we run the prediction over all the training images from target domain. The predicted bounding boxes with high probability scores (0.9 in experiments) will be merged into the original label set $L^s$
to form $L^{s + 1}$.  
We also discard any predicted bounding box 
if it has a high overlap (Intersection-over-Union $>$ 0.2 in experiments) with 
any of the original bounding boxes in $L^s$. 

After $S$ stages, we feed the training images
and the expanded bounding box set $L^{S+1}$
into a normal object detector training pipeline, in which the bounding boxes in 
$L^{S+1}$ are positive while all the rest are negative. 
We still apply the YoloV2 algorithm here for training and testing. 
Ideally, if all the unlabeled object instances could be propagated, the trained model 
should be able to achieve an accuracy on par with that trained from the full annotations.

\section{Experiments}

\subsection{Settings}
\subsubsection{Datasets.}
We mainly evaluate the approaches on the
CARPK~\cite{HsiehLH17} dataset, which contains 989 training images and 459 testing images.
The task is to detect and count the car instances 
in the image.
Each training image has $42.7\pm 15.7$ annotated cars,
while each testing image has $103.5 \pm 39.1$ cars. 
The images are collected by a drone on top of car parks. 
An example image is shown in 
the first row of Fig.~\ref{fig:isl}

Another interesting application is to count the number of drinks or products 
on retail store shelves. 
To demonstrate the effectiveness of the incompletely-supervised learning algorithm on other domains, 
we collect a small dataset \textit{Drink35}, which contains 10 images as the training set and 25 images as the testing set. 
The task is to detect and count all the product instances.
One example image is shown in the second row
of Fig.~\ref{fig:isl}.

To simulate the incompletely-supervised settings, we
randomly select $C_i$ training images, 
and for each image we randomly select at most $C_a$ annotated cars.
All the other unselected images are discarded during training. 
This training set is denoted as 
$\text{CARPK}\_C_i\_C_a$ or $\text{Drink35}\_C_i\_C_a$ for CARPK and Drink35 datasets, respectively. For example, CARPK\_5\_5 means the training set with
5 images and each with at most 5 annotated 
bounding boxes. 
Similarly, the suffix of $\_C_i\_\text{ALL}$
denotes the training set of $C_i$ images with all the annotated boxes, and $\_\text{ALL}\_C_a$ denotes all training images with at most $C_a$ annotated bounding boxes in each image. 
The test set is not altered for consistent evaluation. 

We use the PASCAL VOC~\cite{EveringhamGWWZ10} 2007 trainval set (5011 images) as the extra background images with all the original bounding box labels removed.
Note that the images in PASCAL VOC 2007 contains the object of cars and drinks. 
We keep these images as negative samples because the cars and drinks in VOC 2007 are generally of difference appearances or views compared with the object instances in CARPK and Drink35. 

\subsubsection{Criteria.}
Since we focus on the detection-based approaches, we adopt 
the widely-used~\cite{Girshick15,RenHGS15,LiuAESRFB16,RedmonF17}
mAP@0.5 as one of the metrics, which measures the mean average precision (mAP)
using 0.5 as the interaction-over-union (IoU) threshold. 

Following~\cite{LempitskyZ10,HsiehLH17}, 
we also use the Mean Absolute Error (MAE) and 
the Root Mean Squared Error (RMSE) to evaluate the accuracy of the counting results. 
MAE is defined as $\sum_{i}|n_i - n_i'|/N$, 
while RMSE as $\sum_{i}\sqrt{(n_i - n_i')^2}/N$, 
where $n_i$ is the number of objects predicted by the model for the $i$-th testing image, 
$n_i'$ is the ground-truth number of objects, and $N$ is the total number of testing images. 

\subsubsection{Implementation details.}
The data augmentation is of great importance for the network learning
due to the small training set with incomplete labels. 
Motivated by the implementation of Yolo~\cite{RedmonF17}
and SSD~\cite{LiuAESRFB16},
we incorporate the random scaling,
random aspect ratio distortion, 
and color jittering. 
Random rotation is also implemented for the car counting problem by multiple data samplers (motivated by SSD~\cite{LiuAESRFB16}), 
so that non-rotated images are preferred than rotated images. That is, 
 images with 0, 90, 180, and 270 degrees' rotations are preferred than images with arbitrary  rotations, and images with less than 10 degrees' rotation are preferred than 
images with other arbitrary rotations. 
Specifically, for each training image, we have 25\% chance to 
select non-rotated image; 
25\% chance to rotate images by $90$x degrees; 
25\% chance to rotate images by less than $\pm 10$ degrees plus a random $90$x rotation; and the last 25\% chance to rotate images for any angle. 

Another important parameter is that we use a large input resolution for both training and testing due to large number of objects and small object sizes in each image. During training, we resize the image so that the longer side randomly ranges from $832$ to $1664$, and then crop a subregion of $416\times 416$ as the network input.
During testing, we resize the input image so that the whole area is close to $1248\times 1248$ while its aspect ratio is kept. Non-Maximum Compression (NMS) is used to filter the bounding box and the IoU threshold is 0.2. 

The network backbone is Darknet19, which is the same as in YoloV2~\cite{RedmonF17}.
We use 9 stages of PFOD to propagate the labels. 
In each stage of PFOD, we train the network with 10K iterations. The learning rate is 0.0001 for the first 100 iterations, 0.001 for the next 4900 iterations, 0.0001 for 4000 iterations and 0.00001 for the last 1000 iterations\footnote{The total number iterations could be greatly reduced as the number of training images is small. However, as the training time is not a concern in this work, we leave the training optimization as a future work.}.
The batch size is set to 64, and the weight decay is 0.0005.
The last detector training shares the same parameters. 
The training takes 1.25 hours on 4 NVidia P100 GPUs to finish the 10K iterations. 
The implementation is based on Caffe~\cite{jia2014caffe} under
the environment of CUDA 8.0 and CUDNN 5.1.

We also report the accuracy without any label propagation, and directly train the object detector on the provided label set. This straightforward approach is denoted as OD as a naive baseline,
and all the data augmentation and learning strategy parameters
are the same with a single stage of PFOD.

In the incomplete annotation settings, we enable the extra background samples by replacing 16 images in each batch of 64 with the extra background images. 

\begin{table}[t!]
\centering
\caption{Accuracy on CARPK with different training images.
$C_i$: number of training images from the whole training set; 
$C_a$: number of labeled bounding box annotations for each image;
mAP: the higher, the better; 
MAE/RSME: the lower, the better. 
The value in parenthesis of MAE/RSME is the threshold to decide if the predicted bounding box is valid }
\label{tbl:differentc1c2}
\begin{tabular}{cc@{~~}c@{~~}c@{~~}c@{~~}c}
\toprule
Method & $C_i$ & $C_a$  & mAP(\%) & MAE & RSME \\
\midrule
LPN\cite{HsiehLH17} & 989 & ALL & NotAvail & 23.80 & 36.79 \\
\midrule
\multirow{2}{*}{OD}& 989 & ALL & 96.67 & 2.94 (0.15) & 3.94 (0.10) \\
& $5$ & ALL & 91.97 & 4.38 (0.20) & 5.58 (0.20) \\
\midrule
\multirow{5}{*}{OD} & $5$ & $50$ & 87.90 &  6.02 (0.05)  & 8.23 (0.05) \\
& $5$ & $25$ & 61.26 & 49.28 (0.05)  & 54.45 (0.05) \\
& $5$ & $10$ & 4.95 & 99.39 (0.05)  & 106.51 (0.05) \\
& $5$ & $5$ & 4.58 & 101.93 (0.10) & 108.99 (0.10) \\
\midrule
\multirow{4}{*}{PFOD} & $5$ & $50$ & 83.73 & 11.54 (0.05) & 16.76 (0.05) \\
& $5$ & $25$ & 80.27 & 16.52 (0.05) & 21.45 (0.05) \\
& $5$ & $10$ & 73.46 & 22.47 (0.05) & 27.63 (0.05) \\
& $5$ & $5$ & 72.44 & 23.63 (0.05)  & 26.36 (0.05) \\
\bottomrule
\end{tabular}
\end{table}

\vspace{-0.2cm}
\subsection{Results on CARPK}
The results are shown in Table~\ref{tbl:differentc1c2}
for different numbers of training images ($C_i$) and 
different numbers of labeled bounding boxes ($C_a$) 
in each image.
To count the number of objects, we need a threshold to determine if the predicted
bounding box should be kept from the detector. 
Since different settings might favor different thresholds, we 
select the one with lowest MAE or RSME among $[0.05:0.05:0.7]$.
We select the best threshold for MAE and RSME to examine the best 
performance under these two criteria. 
The threshold is in parentheses of Table~\ref{tbl:differentc1c2}.
Note the criterion of mAP does not depend on the threshold. 
From the table, we have the following observations and discussions. 

\subsubsection{What is the upper bound performance using all the training data?} In the fully supervised setting ($C_i = \text{ALL}$, 
$C_a = \text{ALL}$),
our detector (OD) could achieve 96.67\% mAP@0.5, 2.94 MAE and 3.94 RMSE.  
This significantly outperforms the state-of-the-art method
of LPN~\cite{HsiehLH17}, 
whose MAE is 23.80 and RMSE is 36.79 (mAP is not reported). 
Other baseline approaches are 
not shown here because the accuracy is lower than LPN. 
This demonstrates a strong object detector towards the counting problem. 

\vspace{-0.2cm}
\subsubsection{Do we need hundreds of training images for car counting?} We apply the OD on $C_i=5$
images with all the labeled annotations ($C_a = \text{ALL}$), and the detector can still achieves 91.97 mAP@0.5, 3.00 MAE, and 5.58 RSME without extra background images. Compared with $C_i = \text{ALL}$, $C_a = \text{ALL}$ which uses 989 training images, this result is very encouraging. It clearly indicates that to develop a car counting algorithm, 5 or a few more images might be enough rather than several hundreds.
The 5 image IDs are
20160524\_GF2\_00133, 
20161030\_GF1\_00153, 
20161030\_GF1\_00036, 
20160331\_NTU\_00007,
and \\20161030\_GF2\_00071 for reproducing the result.



\begin{figure}[t]
	\centering
	\begin{tabular}{cc}
		\includegraphics[width=0.48\linewidth]{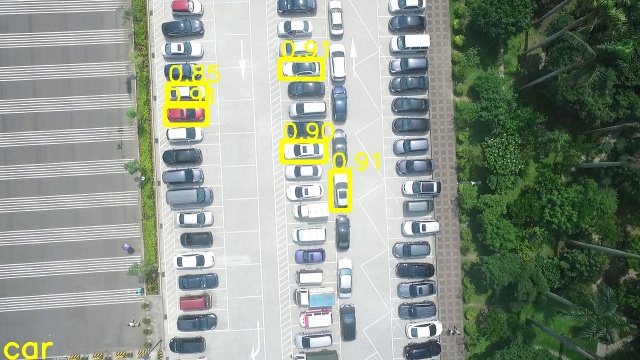} 
        &
		\includegraphics[width=0.48\linewidth]{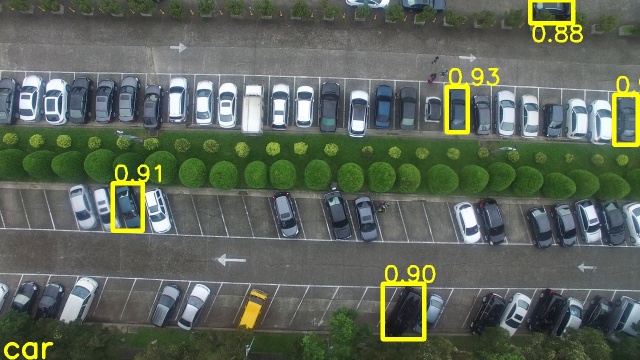} \\
		(a) & (b)
	\end{tabular}
	\caption{Straightforward object detection result on two training images.
 The model is trained on CARPK\_5\_5.
The ground-truth bounding boxes are almost 
identical with the predicted boxes and thus are not shown for clarity. 
Confidence scores shown around each 
box are all close to 1 (0.8+). 
The threshold is 0.05, which means the confidence 
scores for unlabeled/undetected cars are less than 0.05. 
This is a clear indication of overfitting, which severely degrades the performance
}
 \label{fig:overfitting_issue}
\end{figure}
\vspace{-0.2cm}
\subsubsection{What if the unlabeled regions are treated as negative samples?} When the number of bounding boxes is decreased
to 5, the mAP significantly drops to 4.40. 
To identify the reason, we evaluate the trained model 
against the training images and show two examples in Fig \ref{fig:overfitting_issue}. 
The predicted bounding boxes are drawn on the 
image with confidence scores around each box. 
Only the boxes with confidence scores higher than 0.05 are
displayed. The selected 5 labeled boxes are located at the 
same position with the predicted boxes and the probability is 
close to 1. Since the threshold here is 0.05, all
the other regions including the unlabeled cars are classified as background with high confidence. 
This shows that the model overfits the training data severely, 
which degrades the accuracy. 

\begin{figure}[t!]
\begin{tabular}[t]{c@{~}c@{~}c}
\includegraphics[width=\thirdsplitwidth\linewidth]{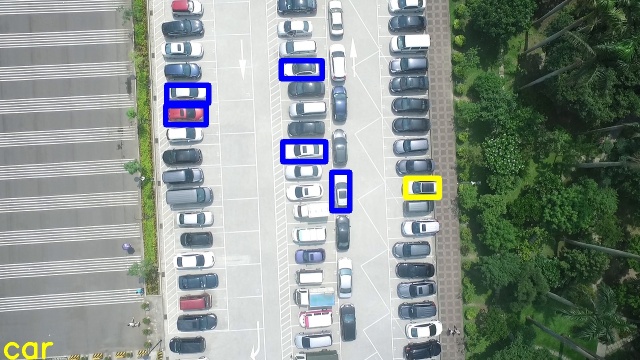} & 
\includegraphics[width=\thirdsplitwidth\linewidth]{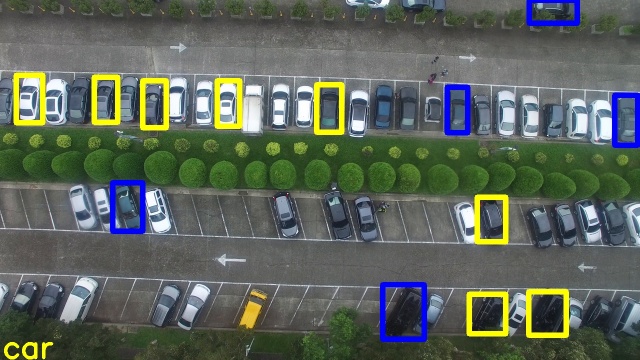} & 
 \includegraphics[width=\thirdsplitwidth\linewidth]{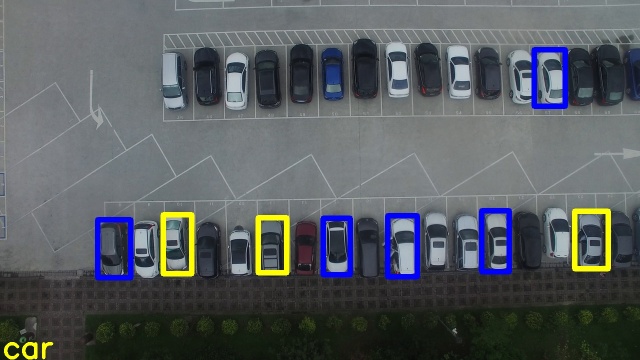} \\
(a) expand=6, correct=6 & (b) expand=13, correct=13 & (c) expand=8, correct=8
\\
  \includegraphics[width=\thirdsplitwidth\linewidth]{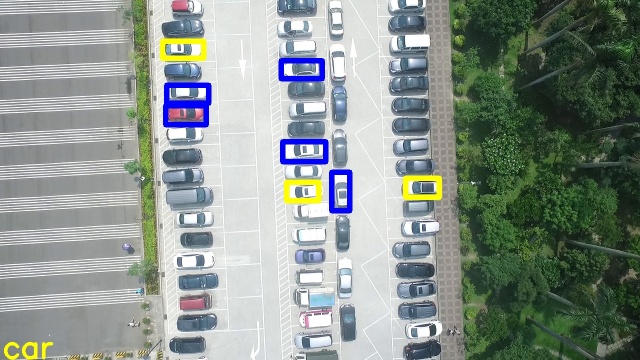} & 
 \includegraphics[width=\thirdsplitwidth\linewidth]{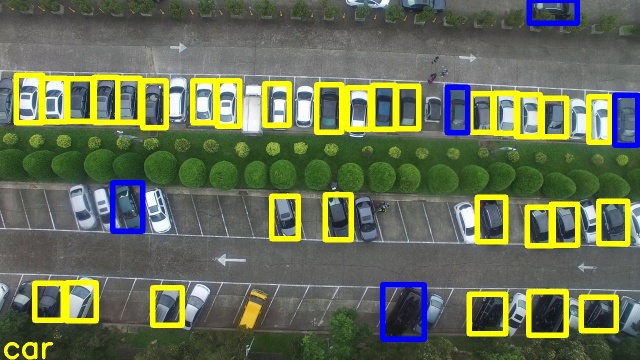} & 
 \includegraphics[width=\thirdsplitwidth\linewidth]{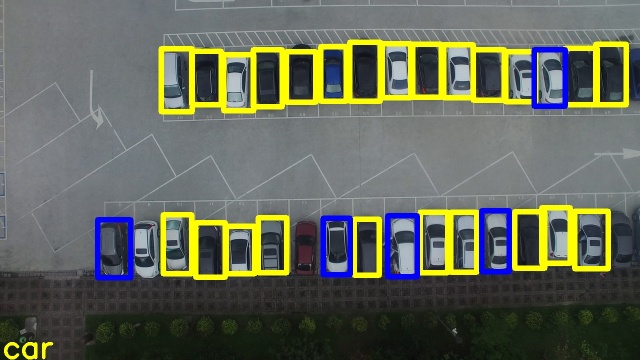} \\
(d) expand=8, correct=8 & (e) expand=35, correct=35 & (f) expand=29, correct=29 \\
  \includegraphics[width=\thirdsplitwidth\linewidth]{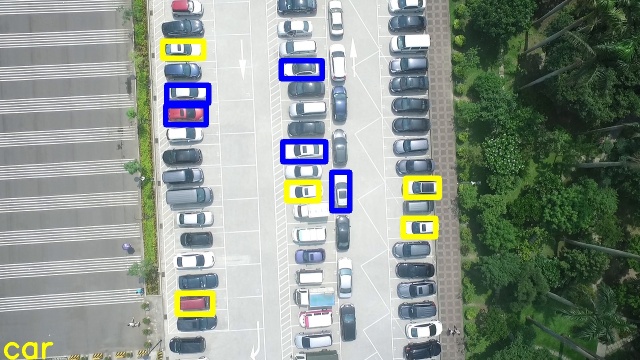} & 
 \includegraphics[width=\thirdsplitwidth\linewidth]{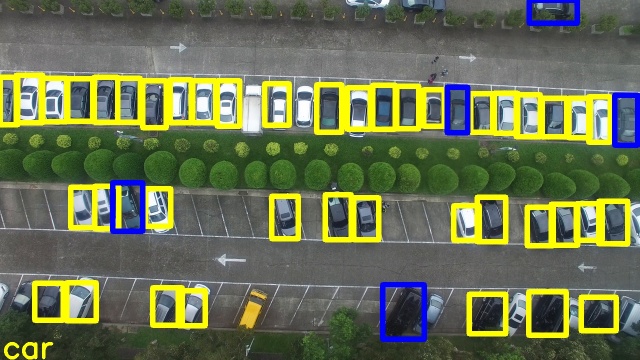} & 
 \includegraphics[width=\thirdsplitwidth\linewidth]{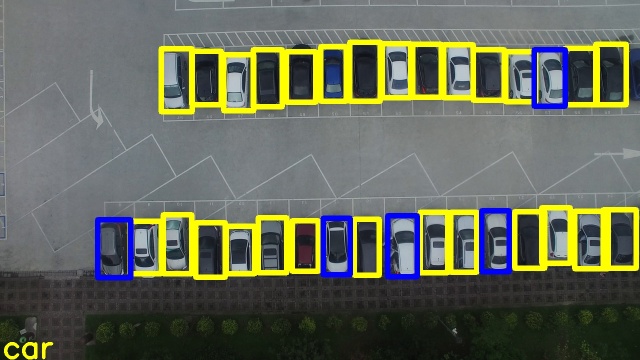} \\
(g) expand=10, correct=10 & (h) expand=46, correct=46 & (i) expand=32, correct=32 \\
   \includegraphics[width=\thirdsplitwidth\linewidth]{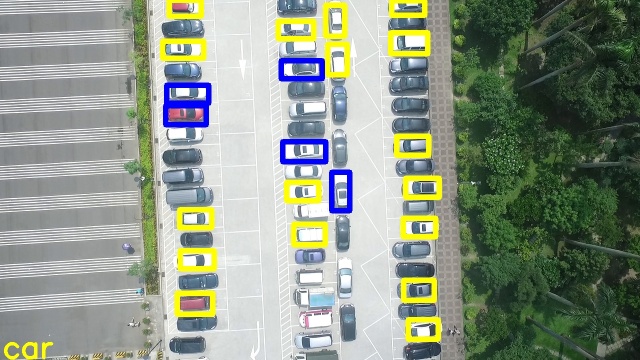} & 
 \includegraphics[width=\thirdsplitwidth\linewidth]{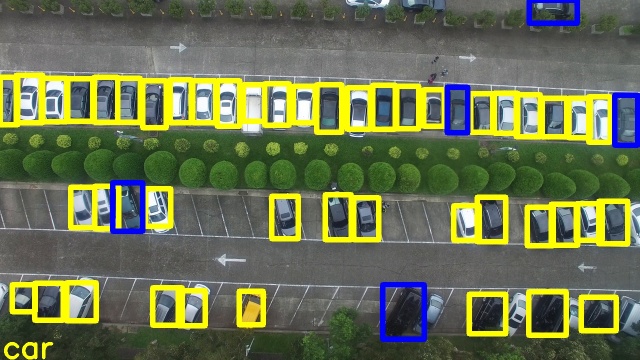} & 
 \includegraphics[width=\thirdsplitwidth\linewidth]{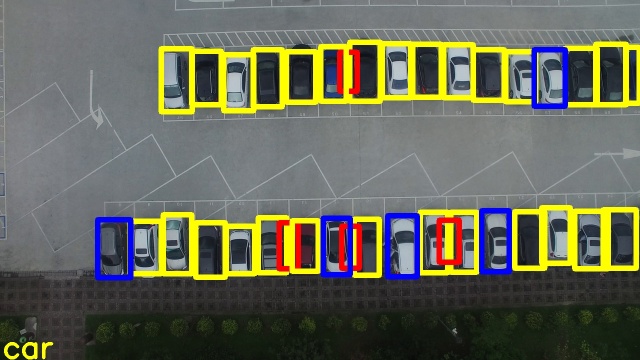} \\
(j) expand=22, correct=22 & (k) expand=50, correct=49 & (l) expand=37, correct=33
\end{tabular}
\caption{Label propagation process on training images of CARPK\_5\_5. 
The two numbers below the images are the number of expanded boxes and the number of correctly expanded. 
From the left to the right, the total number of
true bounding boxes is 61, 58, 34, respectively. 
From the top to bottom, the images correspond to the training data for the third, fifth-th, seven-th and the last stage, respectively. The original sampled 5 labeled bounding boxes are shown in blue; the propagated boxes ares shown in yellow; the incorrectly (IoU $<$ 0.3) propagated is shown in red}
\label{fig:labelpropagation}
\end{figure}

\vspace{-0.2cm}
\subsubsection{How effective is the proposed incompletely-supervised learning approach?} We apply the proposed PFOD 
on CARPK\_5\_5, and surprisingly the accuracy is 
boosted to 72.44 mAP, 23.63 MAE and 26.36 RSME. 
In terms of MAE and RSME, the accuracy has surpassed
the LRN~\cite{HsiehLH17} trained on the full training set of 989 images. 

In Fig.~\ref{fig:labelpropagation}, we illustrate the label propagation process by PFOD on CARPK\_5\_5. Each column corresponds to one training image.
The two numbers below each images are the number of 
the expanded boxes (initially provided + propagated), and the number of correct boxes among those boxes. A box is correct if its IoU
is larger than 0.3 with at least one bounding box
in CARPK\_5\_ALL. 
From the figure, the correct bounding
boxes used for training could 
be gradually populated. Taking the leftmost image as an example, 
the number of correct boxes is increased from the initial number 5 to 22. 
The rightmost one can have 33 correct boxes. 

Meanwhile, we observe that the propagation is still not perfect - it  introduces several false bounding boxes while missing a few cars.
This is the reason why there is still a gap between this setting and CARPK\_5\_ALL, and will motivate us to continue investigating the problem.

\subsubsection{Will introducing more labels help?} By increasing the number of labeled boxes per training image from 5 to 50, 
the accuracy can be smoothly increased for both OD and PFOD. 
With less than 25 boxes in each image, the accuracy of PFOD is consistently 
higher than OD, while with 50 boxes, the accuracy is lower. The reason is 
that under the setting of 50 boxes, most of the true boxes are included, while the label propagation introduced some false boxes.
That is, under the almost full annotations, it is enough to apply the 
OD instead of propagating the boxes. 

\begin{figure}[t!]
\begin{tabular}{c@{~}c@{~}c}
OD on CARPK\_5\_5 & OD on CARPK\_5\_ALL & PFOD on CARPK\_5\_5 \\
\includegraphics[width=\thirdsplitwidth\linewidth]{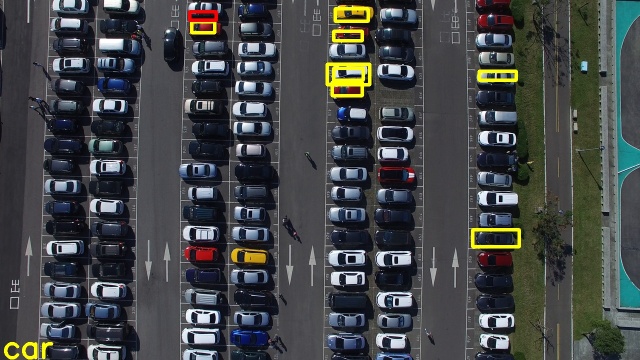} & 
\includegraphics[width=\thirdsplitwidth\linewidth]{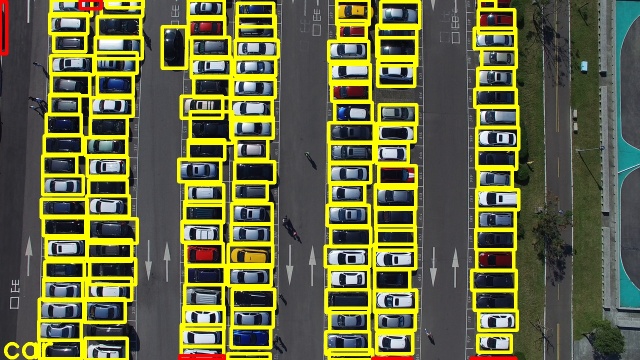} & 
\includegraphics[width=\thirdsplitwidth\linewidth]{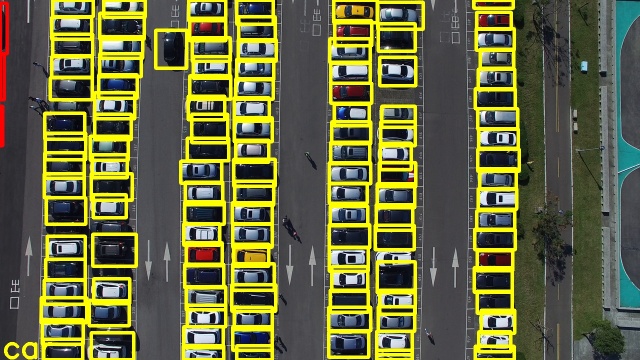} \\
(a) pred=9, gt=127 & (b) pred=130, gt=127 & (c) pred=122, 127 \\
\includegraphics[width=\thirdsplitwidth\linewidth]{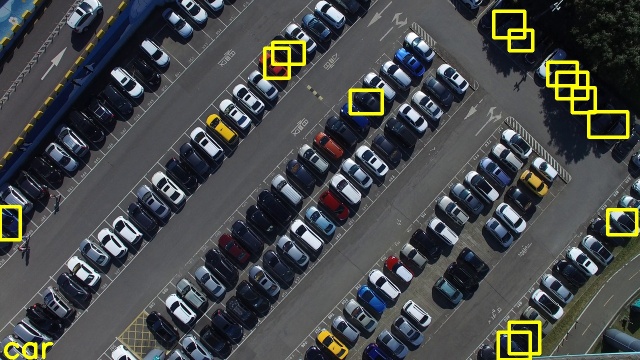} & 
\includegraphics[width=\thirdsplitwidth\linewidth]{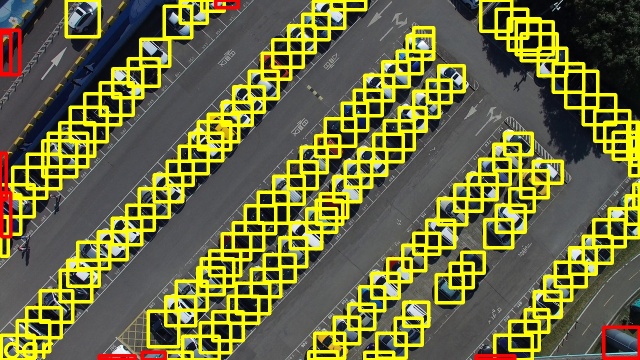} & 
\includegraphics[width=\thirdsplitwidth\linewidth]{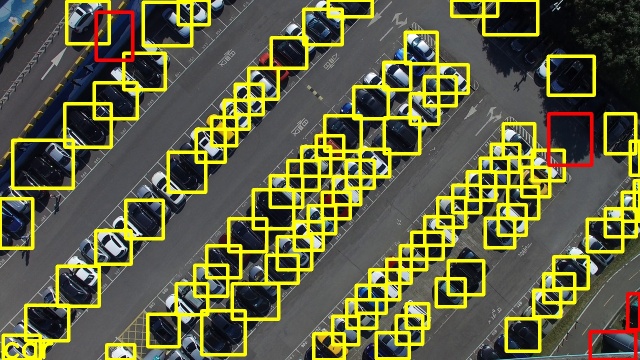} \\
(d) pred=13, gt=138 & (e) pred=152, gt=138 & (f) pred=93, gt=138 \\
\includegraphics[width=\thirdsplitwidth\linewidth]{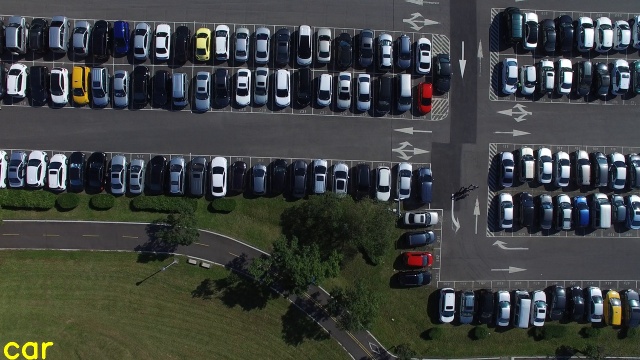} & 
\includegraphics[width=\thirdsplitwidth\linewidth]{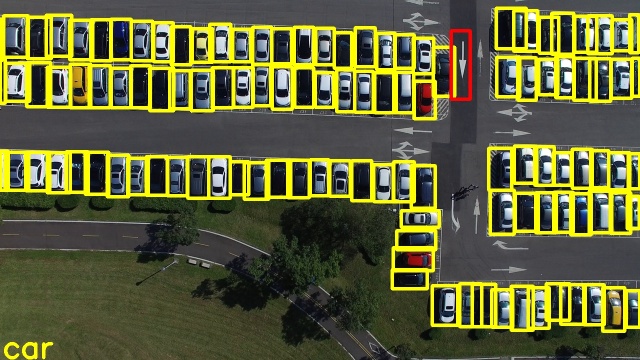} & 
\includegraphics[width=\thirdsplitwidth\linewidth]{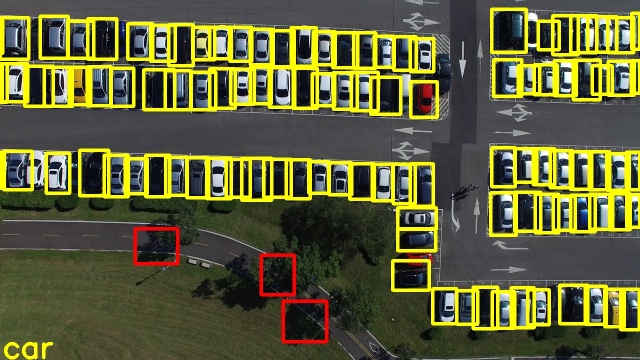} \\
(g) pred=0, gt=114 & (h) pred=115, gt=114 & (i) pred=107, gt=114
\end{tabular}
\caption{Visualization of detection/counting results on 3 CARPK test images. Each row corresponds to one test image. 
Left: OD on CARPK\_5\_5;
Middle: OD on CARPK\_5\_Full; 
Right: our PFOD on CARPK\_5\_5
}
\label{fig:vis_car}
\end{figure}

\begin{figure}[t!]
\begin{tabular}{c@{~}c@{~}c}
OD on Drink35\_ALL\_5 & OD on full Drink35 & PFOD on Drink35\_ALL\_5 \\
\includegraphics[width=\thirdsplitwidth\linewidth]{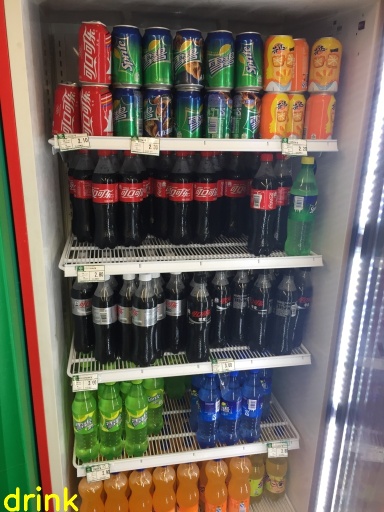} & 
\includegraphics[width=\thirdsplitwidth\linewidth]{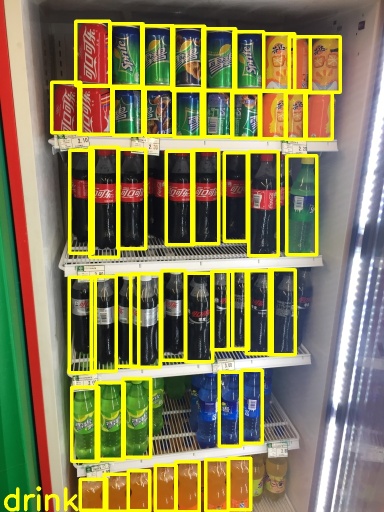} & 
\includegraphics[width=\thirdsplitwidth\linewidth]{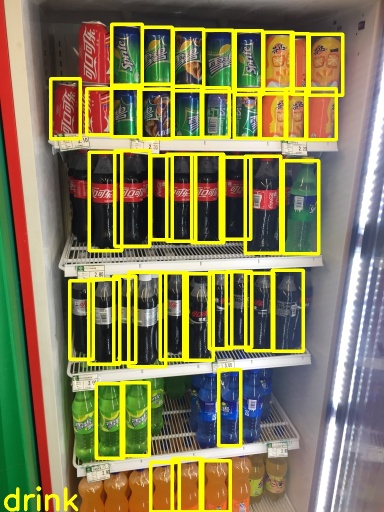} \\
(a) pred=0, gt=58 & (b) pred=49, gt=58 & (c) pred=36, gt=58 \\
  \includegraphics[width=\thirdsplitwidth\linewidth]{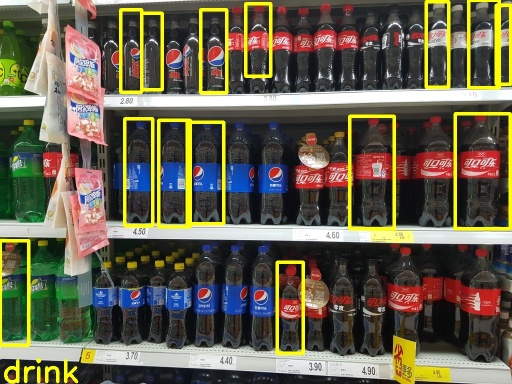} & 
 \includegraphics[width=\thirdsplitwidth\linewidth]{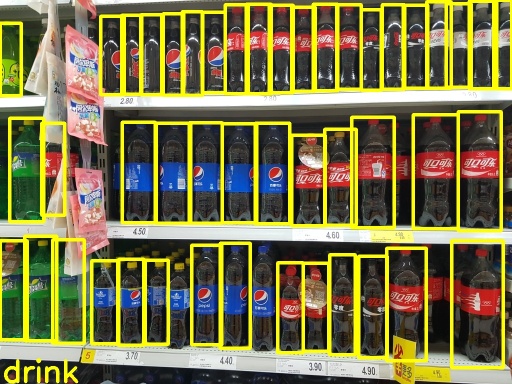} & 
 \includegraphics[width=\thirdsplitwidth\linewidth]{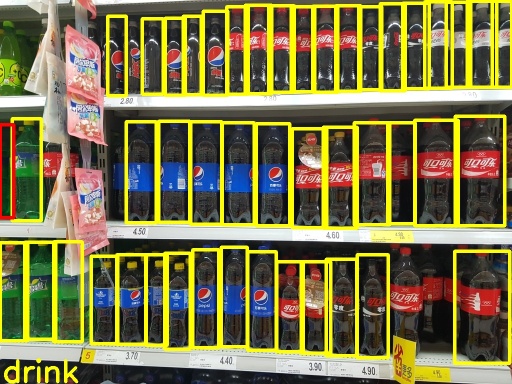} \\
(d) pred=14, gt=51 & (e) pred=46, gt=51 & (f) pred=45, gt=51
\end{tabular}
\caption{Visualization of the detection/counting results on Drink35 test images.
Each row corresponds to one test image. Left: OD on Drink35\_ALL\_5. Middle: OD on the full Drink35; Right: our PFOD on Drink35\_ALL\_5
}
\label{fig:vis_drink}
\end{figure}

Fig.~\ref{fig:vis_car} visualizes the detection and counting results 
on three testing images based on the model with OD trained on CARPK\_5\_5,
OD trained on CARPK\_5\_ALL, and the model with PFOD on CARPK\_5\_5. The yellow boxes are the correct bounding boxes while the red one is the incorrect one. A predicted box is correct if it has an IoU larger than 0.3 with 
one of the ground truth bounding boxes. The two numbers under the image are the number of instances predicted by the model and the ground-truth number of instances, respectively. 

\vspace{-0.2cm}
\subsection{Results on Drink35}
With the full annotations, OD can achieve 79.53\% mAP@0.5, 7.92 MAE and 11.50 RSME. This can be treated as the upper bound performance as we have used all the labels.
If each training image is provided with only 5 labeled instance, 
the OD's accuracy degrades to 14.16\% mAP, 29.52 MAE and 38.62 RSME. In contrast, using PFOD, the accuracy could be jumped to 53.73\%, 11.16 MAE and 13.10 RSME.
%
Fig.~\ref{fig:vis_drink} shows two example images
detected/counted by the three approaches. 

\vspace{-0.2cm}
\section{Conclusion}
We have studied the problem of object counting when there are only a few exemplar annotations available. The problem is more practical especially when the number of object instances is large. We formulate the problem as incompletely-supervised learning in the context of object detection. Since not all the bounding boxes are provided, we cannot simply treat other regions as background which will lead to severe overfitting and performance degradation. To address the problem, we have proposed a positiveness-focused object detector to progressively propagate the incomplete labels to more object instances. Our experimental results over two applications have demonstrated that this simple yet effective approach significantly boosts the accuracy with only a few manually annotations. 


\bibliographystyle{splncs}
\bibliography{main}
\end{document}